\documentclass{dhbenelux}

\usepackage{booktabs} 
\usepackage{authblk} 
\usepackage{graphicx} 

\author[1, 2]{Marco Matarese}
\author[1]{Francesco Rea}
\author[3]{Katharina J. Rohlfing}
\author[1]{Alessandra Sciutti}

\affil[1]{Italian Institute of Technology}
\affil[2]{University of Genoa}
\affil[3]{Paderborn University}

\title{Let people fail! Exploring the influence of explainable virtual and robotic agents in learning-by-doing tasks}

\begin{document}

\maketitle
\copyrightstatement

\begin{abstract}
    \noindent {Collaborative decision-making with artificial intelligence (AI) agents presents opportunities and challenges. While human-AI performance often surpasses that of individuals, the impact of such technology on human behavior remains insufficiently understood, primarily when AI agents can provide justifiable explanations for their suggestions. This study compares the effects of \textit{classic vs. partner-aware} explanations on human behavior and performance during a learning-by-doing task. Three participant groups were involved: one interacting with a computer, another with a humanoid robot, and a third one without assistance. Results indicated that partner-aware explanations influenced participants differently based on the type of artificial agents involved. With the computer, participants enhanced their task completion times. At the same time, those interacting with the humanoid robot were more inclined to follow its suggestions, although they did not reduce their timing. Interestingly, participants autonomously performing the learning-by-doing task demonstrated superior knowledge acquisition than those assisted by explainable AI (XAI). These findings raise profound questions and have significant implications for automated tutoring and human-AI collaboration.}
\end{abstract}

\section{Introduction}
The maturation of artificial intelligence (AI) techniques has facilitated their extensive utilization across various domains. The integration and refinement of explainable AI (XAI) methods have further empowered non-expert users to incorporate AI models into decision-making settings \cite{waldam2022governing}. The resultant dynamics of human-AI collaboration have become a focal point of interest for the human-computer interaction (HCI) community and society at large \cite{bucinca2021trust}.

While human-AI collaboration in decision-making has predominantly been addressed in HCI in recent years, with individuals interacting with artificial agents or receiving suggestions and explanations from recommendation systems \cite{malhi2020explainable, lai2021towards}, the study of the human-robot collaboration received the scientific community's attention since the dawn of the human-robot interaction (HRI) research field. However, recent years have witnessed implementing and testing explainable techniques with robots in collaborative contexts \cite{anjomshoae2019explainable, wallkotter2021explainable}. Differently from the HCI context, the HRI one provides richer interaction modalities, offering a more diverse range of opportunities for personalizing XAI and the modality of the explanations delivery \cite{matarese2021user}. 

An emerging challenge addressed by both the HCI and HRI communities is examining how AI technologies influence human behavior in the context of human-AI collaboration \cite{green2019principles}. Multidisciplinary efforts have investigated the impact of AI suggestions on human decision-making, exploring implications related to human cognitive biases \cite{bertrand2022cognitive}. Moreover, the introduction of XAI techniques has a dual effect, enabling non-expert users to benefit from such powerful technology while also raising concerns about over-reliance on AI models and the reinforcement of negative human heuristics, such as automation bias \cite{vered2023effects}.

This work investigates the impact of interacting with virtual and robotic explainable agents on people's behavior and performance during a learning-by-doing task \cite{anzai1979theory, schank2013learning}. In our experiments, participants had to learn an unknown task with the assistance of an explainable artificial agent, specifically a virtual talking agent and a social humanoid robot. Additionally, a separate group performed the task autonomously without assistance. During the experiments, we employed an assessment task to directly and quantitatively measure the utility of the human-agent explanatory interactions, building on prior work\footnote{M. Matarese, F. Rea, K. Rohlfing, A. Sciutti. How informative is your XAI? Assessing the Quality of Explanations through Information Power (\textit{under review}).}. We aimed to compare the effect of different explanation strategies and explainable agents on participants' behavior, focusing on their final knowledge of the task.

The subsequent sections are organized as follows. Section 2 reviews related works, categorizing them into three parts: human-AI collaboration, explanations in human-AI decision-making, and explanations evaluation. Section 3 outlines the methods employed in the user study, presenting the peculiar methodologies and the technology used during the experiments. Section 4 details the results of the user study, showing comparisons between the experimental conditions. These results are extensively discussed in Section 5, with reference to the existing literature. Finally, Section 6 summarizes our work, highlighting its limitations.

\section{Related Works}

\subsection{Human-AI collaboration}
In the realm of human-AI collaboration, user studies predominantly focus on decision-making \cite{wang2021explanations} or classification tasks \cite{goyal2019counterfactual}. The rationale behind this emphasis lies in the promise of improved performance when assisting human users with expert AI systems \citep{wang2022effects}. While this promise is generally fulfilled, some studies indicate a decline in team performance with certain forms of XAI \citep{schemmer2022meta}. Indeed, the challenge arises from the observation that humans often struggle to detect incorrect AI advice \citep{schemmer2022should, ferreira2021human, janssen2022will}.

Given the potential benefits of collaborating with expert AI partners, non-expert users are particularly interested in AI-assisted decision-making. In a comprehensive review \cite{severes2023human}, authors analyzed the latest XAI literature to validate their insights about non-expert users' interpretation of XAI solutions. Their user study exposed participants to five explanation metaphors to investigate their perception and understanding. Moreover, authors in \cite{lai2021towards} summarized the design choices of 100 papers on decision-making, AI models and AI assistance elements, and evaluation metrics, highlighting current trends and gaps. Scholars have also contributed to the field by systematically reviewing the AI-assisted decision-making literature and proposing a taxonomy of human-AI collaboration interaction \cite{gomez2023designing}.

Scholars have explored approaches such as user awareness and personalization to enhance performance. Recent works on user awareness emphasize trust towards the system, focusing on context-awareness and personalization as crucial for user-centeredness \cite{williams2021towards}. Personalization in XAI has also been implemented by leveraging the users’ personality traits and correlating them with preferences or behaviors \citep{bockle2021can, martijn2022knowing}. Inspired by philosophy and psychology, some works delve into empirical application-specific XAI and the theoretical underpinning of human decision-making. Authors in \cite{wang2019designing} proposed a framework for building human-centered decision-theory-driven XAI, identifying human cognitive patterns that XAI needs to follow and cognitive biases that need to be mitigated. Moreover, they implemented their framework in a clinical diagnostic scenario and drew insights into how their framework bridges XAI and human decision-making theories. Furthermore, authors in \cite{bertrand2022cognitive} reviewed a relevant corpus of literature to understand human biases reflected in XAI methods. 

Trust emerges as a critical issue in AI-assisted decision-making, as people often engage with AI recommendations and explanations without analytical scrutiny, particularly when reinforced by automatic explanations. Strategies proposed to foster appropriate trust include considering both AI and human confidence \cite{ma2023who}, suggesting that decision-makers need to understand when to trust AI and when to trust themselves. They showed that also considering humans' confidence promoted appropriate trust in AI compared to only using the AI confidence scores. Researchers have also demonstrated that cognitive forcing mitigates humans' over-reliance \cite{danry2020wereable}. For instance, authors in \cite{bucinca2021trust} designed three cognitive forcing strategies to engage more thoughtfully with AI-generated explanations, demonstrating a reduction in over-reliance compared to simple XAI approaches.

\subsection{Explanations in collaborative decision making}
The investigation of the impact of XAI techniques on human users in decision-making tasks is predominantly situated in the human-computer interaction (HCI) field \cite{gambino2022considering, lai2021towards}. Such works often include comparisons between different types of explanations. For instance, authors in \cite{lim2009why} explored the effects of why and why not-explanations on users' system understanding, revealing that why-explanations fostered better understanding and trust towards the system compared to why-not-explanations. Recent efforts have delved into finer comparisons, such as rule- vs. example-based explanations with decision support systems, assessing their impact on system understanding and persuasiveness \citep{vanderwaar2021evaluating}. 

Consequently, user-centredness and personalization have been driving factors in XAI-related research in HCI. In \cite{millecamp2019explain}, authors showed that users' characteristics, such as the need for cognition, influence the interaction with explainable recommendation systems. However, a subsequent study \cite{millecamp2020what} identified that users' openness plays a role in determining their willingness to reuse the explanatory systems. Similarly, in \cite{conati2021toward}, the authors demonstrated that providing explanations increases users' trust and perceived usefulness, offering insights on personalizing explanations based on users' personality traits. Furthermore, researchers evaluated personalized explanations, showing higher user satisfaction compared to non-personalized ones \cite{tintarev2012evaluating}.

In the context of human-robot interaction (HRI) context, there is a scarcity of studies on XAI with decision-making tasks, especially those investigating different XAI strategies or customization and user-centredness \cite{setchi2020explainable}. Notable exceptions include works like \cite{kaptein2017personalised}, where authors explored preference for explanation styles in a belief-desire-intention agent on a Nao robot among both children and adults. Adults preferred goal-based explanations, while children did not exhibit specific preferences. Previous research has also been conducted regarding the persuasiveness of explainable robots during collaboration with humans \cite{matarese2023ex, matarese2023natural}. Specifically, these highlighted that explanations based on the human-robot common ground can influence people more than accurate ones \cite{matarese2023ex} and that people's personality dimensions contribute to their willingness to accept the robot's justified suggestions \cite{matarese2023natural}.

While few studies focus on XAI with decision-making tasks in HRI, several approaches have been proposed to explain robot planning. For instance, authors in \cite{chakraborti2017plan} proposed to afford explainability as a reconciliation model using the Fetch robot. Their approach aims to progressively change the human model to bring it closer to the robot's, making the robot's plan optimal for such changes in the human model. Moreover, authors in \cite{sukkerd2018towards} proposed an explainable planning representation to ease explanation generation and a method to generate contrastive explanations as policy justification. Finally, \cite{devin2016implemented} moved towards the direction of a user-aware XAI during shared plan execution.

Virtual embodied agents have also been the subject of various studies in this regard \citep{anjomshoae2019explainable}. Researchers explained robot behavior as intention signaling using natural language sentences, evaluating their approach through an online study with a virtual robot \cite{gong2018behavior}. Others introduced augmented reality to display XAI feedback and the robot's internal beliefs, testing their method with a virtual Johnny robot and demonstrating its potential to enhance the HRI \cite{wang2022investigating}. Differently, authors in \cite{amir2018highlights} developed an algorithm to summarize robots' behaviors using information extracted from the agent simulations, evaluating their algorithm with a computer in a game-like scenario.

Robots' influence and persuasiveness have been studied for a long time since researchers observed that several influence mechanisms that occur between humans also occur in HRI \cite{saunderson2019robots}. Authors in \cite{hashemian2019power} explored persuasive robot strategies based on social power, presenting information or providing social rewards. Similarly, \cite{saunderson2022investigating} investigated how a robot’s persuasive behavior influences people’s decision-making in a guessing game, comparing emotional versus logical strategies. Moreover, coming to the XAI field, researchers have taken the study on trust and influence with explainable robots to the extremes, investigating the effect of an emergency (simulated) robot's explanations during an evacuation when differing from the crowd movements \cite{nayyar2020exploring}.

This work contributes by investigating learning during the interaction with an explainable robot, measuring the degree of alignment with its behavior before receiving its suggestions. While limited prior research has explored learning effects with explainable robots, numerous studies have focused on XAI teaching and the impact of explanations on users' learning in HCI \cite{krzysztof2022explainable}. 
Recent HCI works proposed practices and methods for incorporating XAI in teaching \citep{clancey2021methods}, highlighting benefits in supporting trust calibration and enabling rich teaching feedback forms \citep{ghai2021explainable}. Authors in \cite{simkute2022xai} delineated a strategy, based on cognitive psychology literature, for tailoring XAI interface design to have a long-lasting educational impact. Furthermore, authors in  \cite{khosravi2022explainable} defined a framework for XAI in education, considering key aspects to study, design, and develop educational AI tools.

\subsection{Evaluating the quality of explanations}
Despite the increasing number of works focusing on the quality of XAI, defining what constitutes a good explanation remains a challenge. Researchers in the field acknowledge the urgency of establishing metrics for XAI \citep{nauta2022anecdotal}, yet defining objective metrics is challenging due to the strong dependence of XAI model efficacy on the application context and users' expertise \citep{hoffman2018metrics}. 
Consequently, recent efforts within the community have concentrated on elucidating the quality of explanations. For instance, authors in \cite{wang2021explanations} outlined three desiderata for XAI systems: the ability to (1) understand the AI model, (2) recognize the uncertainty underlying an AI prediction, and (3) calibrate their trust in the model. 

Given that explanations are intended for human users, evaluations of XAI systems' properties are often conducted through user studies, with a particular emphasis on non-expert users \cite{janssen2022will}. These studies often involve tasks unfamiliar to most people, ranging from ordinary tasks such as optimizing insulin doses \cite{van2021evaluating} or predicting recidivism \cite{wang2021explanations} to more unconventional scenarios like assessing alien food preferences \cite{lage2019evaluation} or classifying bird species \cite{goyal2019counterfactual, wang2020scout}.

While many works assessing or comparing XAI methods tend to define their own measures of goodness \citep{van2021evaluating, lage2019evaluation},  a method has been proposed to objectively measure the \textit{degree of explainability} of information provided by an XAI system \citep{sovrano2022quantify}. This method quantitatively measures how many \textit{archetypical} questions the system can answer. Moreover, authors in \cite{holzinger2020measuring} proposed the \textit{System Causability Scale} to measure the explanations' quality based on causability \citep{holzinger2019causability}. Taking a different perspective, authors in \cite{wang2022effects} compared different types of explanations in various application contexts, assessing their effectiveness based on three desiderata: improving people’s understanding of the AI model, helping in recognizing the model uncertainty, and supporting calibrated trust in the model.

\begin{figure}[t]
    \centering
    \includegraphics[width=\linewidth]{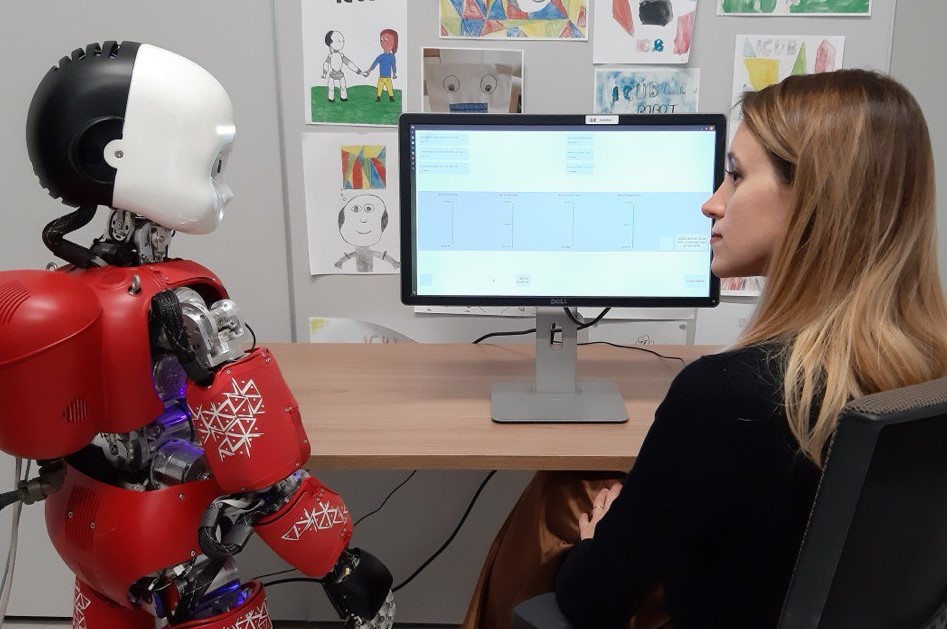}
    \caption{A participant interacts with the humanoid robot iCub during the training phase (Robot group). On the screen is the nuclear power plant application running.}
    \label{fig:setup_NPP}
\end{figure}

\section{Material and Methods}
\begin{figure}[t]
    \centering
    \includegraphics[width=.8\linewidth]{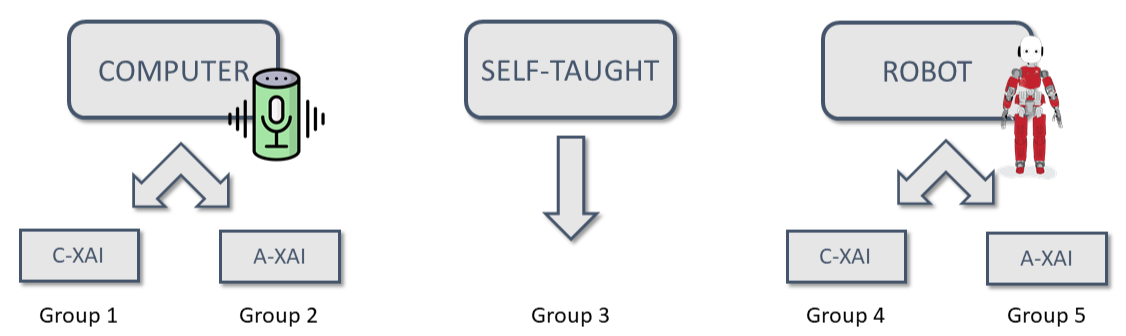}
    \caption{Experimental design for the NPP experiment. Three macro-groups were considered: COM (with the computer), Self-taught, and Robot. Participants from the COM and Robot groups were divided into two experimental conditions (C-XAI and A-XAI) depending on the explanation strategy adopted by the agent.}
    \label{fig:design_npp}
\end{figure}

This study aims to investigate the informativeness of contrastive explanations (A-XAI) compared to classical causal explanations (C-XAI) during a learning-by-doing task and explore the impact of the explainable agent type on participants' behavior. To this aim, we conducted a between-subject user study. One group of participants (22 in total) interacted with a computer (the COM group), while another group (22 in total) engaged with a social humanoid robot (the Robot group) like in Figure \ref{fig:setup_NPP}. Both groups were further divided based on the type of explanation experienced during the task. Furthermore, we conducted the same experiment with a baseline group of participants (11) who worked autonomously without interacting with any agents (the Self-taught group). Figure \ref{fig:design_npp} provides an overview of the group distribution.

Before starting the experiment, all participants signed the informed consent. The informed consent that the University approved participants from the COM group signed of Paderborn ethical committee because such experiments have been performed in Paderborn, Germany. The remaining participants signed an informed consent approved by the ethical committee of ``Regione Liguria'' because the experiments were carried out in Genoa, Italy.

\subsection{The task}
\label{sec:task_npp}
We considered non-expert users to ensure the information passed through the interaction was new. Thus, we considered only participants without knowledge about the task and its underlying rules. We implemented a nuclear power plant (NPP) management task, chosen for several reasons: its challenging yet engaging nature for non-expert users, simple governing rules, the potential for an AI model to learn these rules, and the typical lack of knowledge people have about the functioning of nuclear power plants.

The main objectives of the task (which we hid from users) were to generate maximum energy and maintain the system in an equilibrium state. The features of the environment were subject to rules and constraints summarized as follows:
\begin{itemize}
    \item Each action triggered a change in the environment's features, affecting its values.
    \item Specific preconditions had to be met to start and sustain nuclear fission for energy production.
    \item Some conditions irremediably damage the NPP.
\end{itemize}

The assessment involved a decision-making learning-by-doing task where users interacted with a control panel to perform actions in a simulated environment. During the task, users could engage with an expert explainable AI agent by pressing buttons on the application GUI to ask \textit{what} it would do in each step. Moreover, they could also ask \textit{why} it would perform a specific action. To both these questions, the agents replied verbally. Aside from instructions regarding interacting with the control panel and the agent, users start the task with no prior knowledge.

In a fixed time frame of 30 minutes, the users had to discover:
\begin{itemize}
    \item The nature of the task (\textit{e.g.}, its goals).
    \item The rules governing the simulated environment.
    \item The rules guiding the AI model's actions.
\end{itemize}

Participants could interact with the environment by pressing buttons and setting sliders (Figure \ref{fig:setup_NPP}) to perform actions and observe the outcomes. Moreover, they could seek additional information from the artificial agent through its suggestions and explanations. Following this learning phase, participants have to complete an \textit{assessment} phase. In this stage, participants had to perform the task they learned during training, aiming to achieve the best performance within a fixed time frame of 10 minutes. Subsequently, users underwent a test evaluating their comprehension of the task's objectives and rules. This test could take various forms, such as open-ended or multiple-choice questions. We opted for both types of questions and aimed to assess:
\begin{itemize}
    \item Users' knowledge of the task's objectives.
    \item Users' understanding of the task's internal rules.
    \item Users' ability to generalize the skills acquired during the task.
\end{itemize} 

The roles within the human-robot/computer teams and their interaction modalities were predefined. The artificial agents were limited to assisting users during decision-making and could not perform actions themselves. Moreover, the agents could not provide suggestions on their own initiative; they could only respond to explicit user queries. Consequently, only the user interacted with the control panel and acted within the simulated environment. For the Robot group, iCub and participants sat side-by-side to improve their attitudes toward the robot during the collaboration \citep{geiskkovitch2020where}.

\subsection{The simulated environment}
\label{sec:env_npp}
To design the functioning of our simulation, we started with the NPP's actual functioning and simplified it to reach a trade-off between complexity and feasibility. Our simulated power plant comprised four continuous features: reactor core pressure, the temperature of the water in the reactor, water amount in the steam generator, and reactor power. Furthermore, the power plant had four other discrete features related to the reactor rods: security rods, fuel rods, sustain rods, and regulatory rods. The first two had two levels: up and down. Instead, the latter two had three levels: up, medium, and down.

The reactor power linearly decreased over time due to the de-potentiation of the fuel rods. Hence, reactor power depends on the environment's features values and the occurrence of nuclear fission. The energy produced at each step is calculated by dividing the reactor power by 360, representing the power produced by a 1000MW reactor without power dispersion in 10 seconds (the expected duration of participant actions).

Users could perform 12 actions, ranging from adjusting rods' positions to adding water to the steam generator or skipping to the next step. Each action modified the value of three parameters, corresponding to the water's temperature in the core, core pressure, and the water level in the steam generator. 

The rod configurations determined the magnitude of feature updates performed at the end of each step after the users' action. For instance, lowering the safety rods halted nuclear fission, leading to a decrease in the core temperature and pressure (unless they reach their initial values), with no changes in the water level of the steam generator. Conversely, if nuclear fission occurred and the user lowered the regulatory rods, the fission accelerated. This acceleration consumed more water in the steam generator, raising the core temperature, pressure, reactor power, and electricity faster than in normal conditions.

\subsection{The agents' AI}
\label{sec:ai_npp}
Regarding the agents' AI model, we employed a deterministic decision tree (DT) trained using the Conservative Q-Improvement (CQI) learning algorithm \citep{roth2019conservative}. This approach allowed us to train the DT using a reinforcement learning (RL) strategy, avoiding the extraction of the DT from a more complex ML model \citep{vasilev2020decision, xiong2017learning}. CQI learns a policy as a DT by splitting its current nodes only if it represents a policy improvement. Leaf nodes correspond to abstract states and indicate the action to be taken, while branch nodes have two children and a splitting condition based on a feature of the state space. Over time, their algorithm creates branches by replacing existing leaf nodes if the final result represents an improved policy. In this sense, the algorithm is considered additive, while it is conservative in performing the splits \citep{roth2019conservative}.

We used RL for two main reasons: (1) we did not have data to train our ML model, and (2) we could better control the model behavior by designing a proper reward function. Indeed, we defined the RL reward to maximize the amount of energy produced without damaging the NPP and prioritizing actions having effects on the environment's features (e.g., actually changing their values). We obtained a DT that exhibited solid behavior and optimal performance on our simulated NPP.

Instead of extracting the DT from a more complex ML model \citep{vasilev2020decision, xiong2017learning}, we used this learning strategy to simplify the translation from the AI to the XAI. Moreover, using a binary DT, we obtained an explainable AI model without sacrificing performance. There is a broad consensus on the inherent transparency of such models, even though some authors questioned the assumption that simpler models are more interpretable than complex ones \cite{furnkranz2020cognitive}. The artificial agents used this expert DT to choose their action among the twelve possible actions based on the eight environment's features. 

Starting from its root node, the DT was queried on each internal node, representing binary splits, to determine which sub-trees continue the descent. Each internal node regarded a feature $x_i$ and a value for that feature $v_i$: the left sub-tree contained instances with values of $x_i \leq v_i$, while the right sub-tree contained instances with values of $x_i > v_i$ \citep{buhrman2002complexity}.

The DT's leaf nodes represented actions, defined with an array containing the actions' expected Q-values in implementing \cite{roth2019conservative}. The higher Q-value was associated with the most valuable action. This design enabled the DT to respond to users' both what and why questions. To address a what question, we only needed to navigate the DT using the current values of the environment's features and present the resulting action to the user. To answer a why question, we could provide one of the features' values encountered during the descent.

\begin{figure}[t]
    \centering
    \includegraphics[width=.8\linewidth]{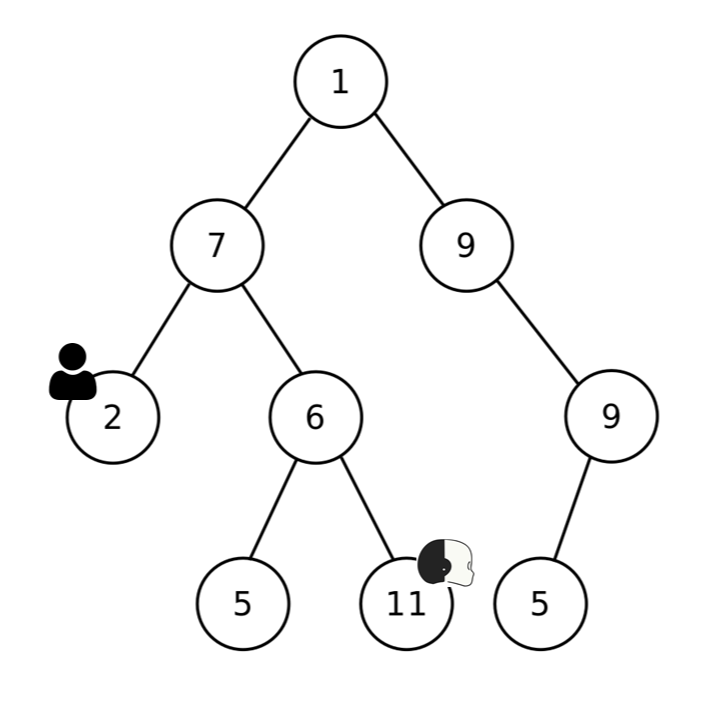}
    \caption{Example of DT where the leaf nodes 11 and 2 are the robot's suggestion and the predicted user's action, respectively. The \textit{classical} XAI selects node 1 for the explanations since it is the most unused relevant node. The \textit{partner-aware} XAI selects node 7 instead because it represents a perfect contrastive explanation for the fact 11 and foil 2.}
    \label{fig:dt}
\end{figure}

\subsection{The agents' XAI}
Since the AI model to explain is inherently transparent \citep{adadi2018peeking}, we can directly exploit the DT to provide explanations by using one of the feature values we encounter during the tree's descent.

As explained in Section \ref{sec:ai_npp}, during the DT descent, we encounter a set of split nodes defined by a feature $x_i$ and a value $v_i$; the direction of the descent tells us whether the current scenario has a value of $x_i \leq v_i$ or $x_i > v_i$. Each of those inequalities can be used to provide an explanation, aiding users in relating actions with specific values of the environment's features. In our case, an explanation for the action ``add water to the steam generator'' could be ``because the water level in the steam generator is $\leq 25$'' (dangerously low).

The problem of selecting which feature to use among those encountered during descent is referred to as \textit{explanation selection}. In our case, we compared two explanation selection strategies. Classical approaches typically use the most relevant features (based on measures such as the Gini index, information gain, or other established metrics \citep{stoffel2001selecting}). We planned to compare a classical explanation strategy with a contrastive, partner-aware one.

The \textit{classical} XAI explains using only the AI outcomes and environment states. In particular, it justifies the agent's suggestions using the most relevant features, which were the first ones in the DT's structure (see \citep{roth2019conservative}). However, the system tried to provide different explanations in each step. It did so by keeping track of the DT's nodes already used and preferring to use the not-used ones by descending the DT's structure.

On the other hand, the \textit{partner-aware} XAI approach, through monitoring and scaffolding, took into consideration the partner's action indication and used such indications to provide contrastive explanations. The facts (the outcome to explain) were the agents' suggestions, while the foils (the expected outcome) were the predicted users' actions. Figure \ref{fig:dt} illustrates an example of these two approaches.

We categorized our findings based on whether they focused on the influence of the computer, the social robot, or self-learning. We expected that:
\begin{itemize}
    \item H1: adaptive explanations would elicit more accurate participants' mental models about the task because we expected that the contrastive nature of such explanations would help participants understand the task's rules better and faster.
    \item H2: regarding participants' learning, the robot, and the computer would produce comparable outcomes, while assisted participants would outperform the not-assisted ones.
    \item H3: participants interacting with the robot would be more persuaded by it, with a higher impact of their personality dimensions on their decision-making due to the social interaction.
\end{itemize}

\begin{figure}[t]
    \centering
    \includegraphics[width=.8\linewidth]{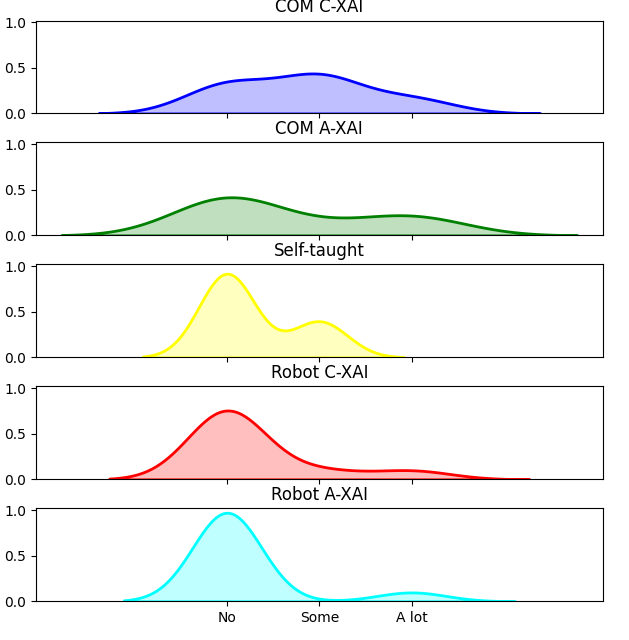}
    \caption{Distribution of the participant's knowledge about the functioning of nuclear power plants before the experiment. We classified them into three levels of knowledge (No, Some, A lot) by coding their open-ended questions to the pre-experiment questionnaire.}
    \label{fig:knowledge_density}
\end{figure}

\section{Results}

\subsection{A priori knowledge of the task}
Since we used a between-subject approach, we needed to ensure that the different groups had comparable starting conditions, \textit{i.e.} that they started with similar task knowledge. To measure participants' knowledge about the functioning of NPPs, we asked them to describe such functioning in an open-ended question during the pre-experiment questionnaire. Two student colleagues then coded their answers in three levels of understanding (No, Some, and A lot of knowledge) based on whether they reported incorrect or no information, some information, or correct information, respectively. Subsequently, we performed a $\chi^2$ test to investigate the differences in the distribution of participants' prior levels of understanding among the experimental groups. We found no significant differences among such distributions ($\chi^2$ test: $\chi^2(8) = 14.1$, $p = .078$).
The distribution of participants' level of prior knowledge is shown in Figure \ref{fig:knowledge_density}.

\subsection{Quantitative differences between classical and adaptive explanations}
As mentioned, we manipulated our experiments on the type of explanations the artificial agents provided. For one group of participants, we provided \textit{classical} explanations (C-XAI), while for the other group, we provided contrastive \textit{adaptive} explanations (A-XAI). Since the difference between the two explanation strategies resided in the explanation selection, we counted (in percentage) the number of explanation topics (the explanandum) used by the artificial agent during the training phase. 

\begin{table*}[t]
    \centering
    \begin{tabular}{lcccccccc}
    \toprule
                 & \multicolumn{4}{c}{\textbf{COM}}                         & \multicolumn{4}{c}{\textbf{Robot}}                       \\ 
                 \cmidrule(lr){2-5}\cmidrule(lr){6-9}
                \textbf{Explanandum} & \textit{\textbf{t}} & \textit{\textbf{p}}      & \textit{\textbf{$\mu_{C}$}} & \textit{\textbf{$\mu_{A}$}} & \textit{\textbf{t}} & \textit{\textbf{p}}      & \textit{\textbf{$\mu_{C}$}} & \textit{\textbf{$\mu_{A}$}} \\ \hline
Temperature      & 50.86      & \textless .001* & 22\%        & 7\%         & 3.76       & \textless .001* & 24\%        & 15\%        \\
Pressure         & 11.58      & .001*           & 28\%        & 14\%        & 3.74       & \textless .001* & 22\%        & 10\%        \\
Water in the steam generator & 13.43      & \textless .001* & 19\%        & 30\%        & .81        & .42             & 24\%        & 27\%        \\
Power            & 1.46       & .23             & 16\%        & 17\%        & 1.17       & .24             & 17\%        & 18\%        \\
Safety rods      & 13.87      & \textless .001* & 7\%         & 18\%        & -2.2       & .03*            & 5\%         & 14\%        \\
Regulatory rods  & 10.45      & .002*           & 1\%         & 3\%         & .06        & .94             & 0\%         & 0\%         \\
Sustain rods     & .77        & .38             & 8\%         & 9\%         & -1.8       & .04*            & 8\%         & 15\%        \\
Fuel rods        & 2.92       & .09             & 1\%         & 2\%         & .1         & .91             & 1\%         & 1\%         \\ \hline
\end{tabular}
    \caption{Comparison of the occurrences (in percentage) of the explanandum between the experimental conditions: we signed with an * those comparisons that were significantly different (independent samples t-test).}
    \label{tab:explanandums}
\end{table*}

We compared such percentages among the experimental groups to check whether different explanation strategies brought different explanandum. We performed an independent samples t-test on each possible explanandum (the environment's features) between the C-XAI and A-XAI groups for both the COM and Robot groups. Regarding the COM group, we found significant differences between explanandum percentages in five out of eight comparisons. Instead, regarding the Robot group, we found significant differences in four out of eight comparisons. Table \ref{tab:explanandums} shows that our manipulation occurred by reporting all the comparisons with their statistics.

Moreover, we investigated whether there were also differences in the question-asking frequencies between the conditions. Hence, we performed a $\chi^2$ test on the questions' frequencies. Such tests showed no significant differences between the groups about both the what- and why-questions.

\subsection{The influence of the computer}

\begin{figure}
    \centering
    \includegraphics[width=.8\linewidth]{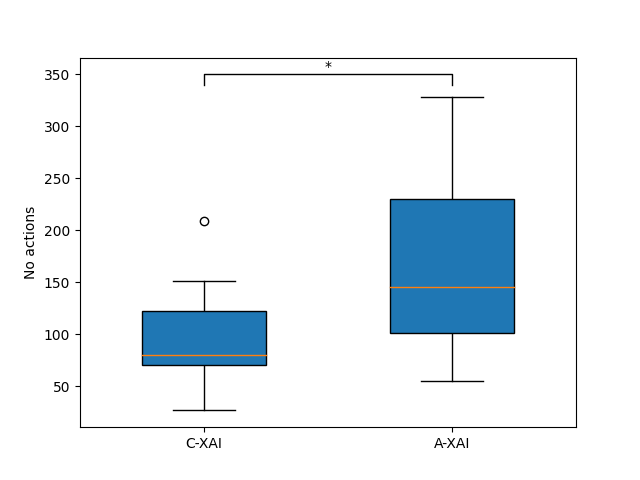}
    \caption{The number of actions participants in the COM group performed during training. The * represents a statistically significant difference  (\textit{p-value} = .039, independent samples t-test).}
    \label{fig:actions_com}
\end{figure}

We found that participants who received adaptive explanations performed significantly more actions than those who received classical explanations (independent samples t-test: $t = 2.21$, $p = .039$). Figure \ref{fig:actions_com} shows a box plot of participants' actions during the training phase. However, we did not find differences in the behavioral measures regarding the assessment phase between the two groups.

\begin{figure*}[t]
    \centering
    \includegraphics[width=.49\linewidth]{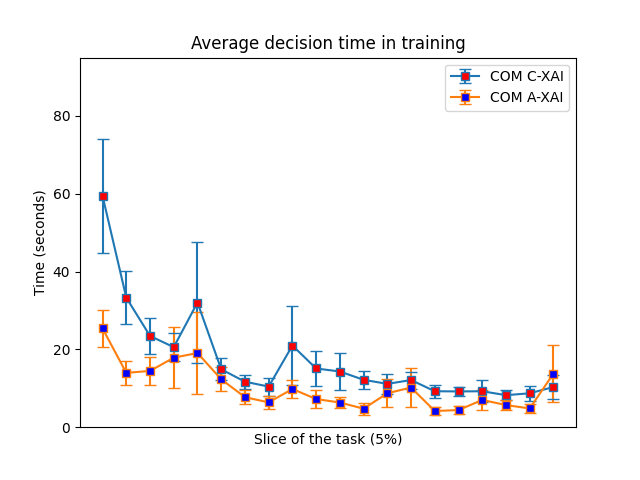}
    \includegraphics[width=.49\linewidth]{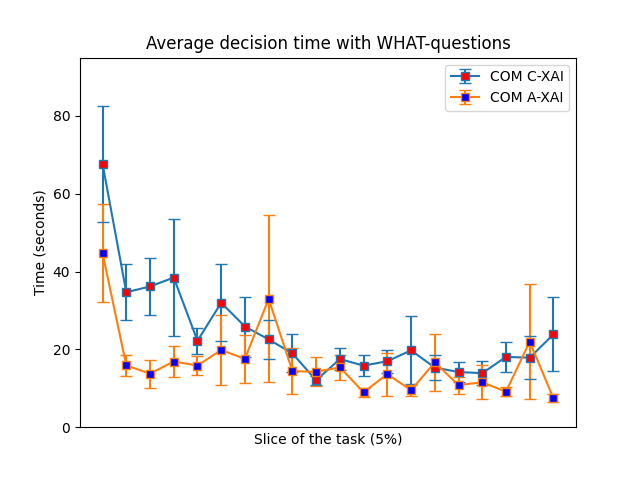}
    \includegraphics[width=.49\linewidth]{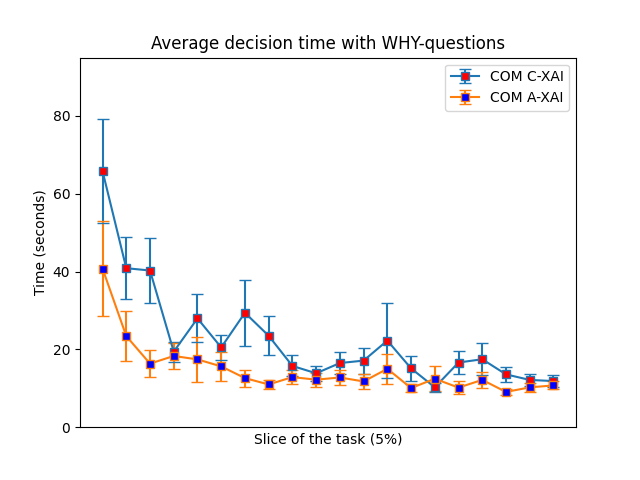}
    \includegraphics[width=.49\linewidth]{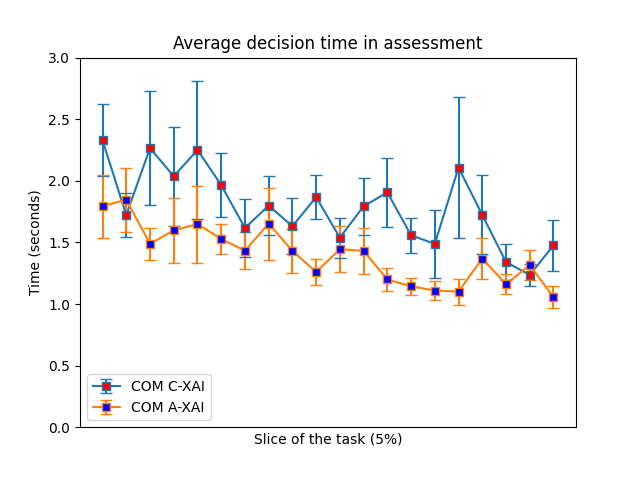}
    \caption{Average and std error of participants' decision times (COM group). Each point of the x-axis represents 5\% of the training phase: it has to be read from the left to the right. The plot on the up-left side shows the average decision time of all steps; the one on the up-right side regarded those where participants asked what questions; the plot on the bottom-left side regarded those where participants asked why questions. The one on the bottom-right side shows the assessment decision time.}
    \label{fig:dt_com}
\end{figure*}

This result is explained by the significant difference between the average moving time of the two groups (independent samples t-test: $t = 2.3$, $p = .02$), as shown in Figure \ref{fig:dt_com}. We also found that the decision time of the steps in which participants asked what- and why questions were significantly different between the C-XAI and A-XAI groups (independent samples t-test: $t = 2.19$, $p = .03$, and $t = 2.3$, $p = .02$, respectively). Finally, we found that the decision times during the assessment phase differed significantly between the two groups (independent samples t-test: $t = 4.4$, $p < .001$), with the A-XAI group moving significantly faster than the C-XAI one.

\begin{figure}[h]
    \centering
    \includegraphics[width=.8\linewidth]{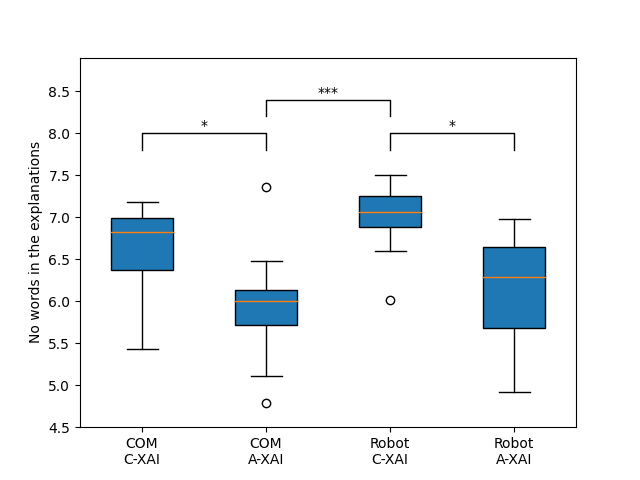}
    \caption{Explanations' length measured as the number of words used by the artificial agents. The * refers to a significant difference with $p-value < .5$, and the *** refers to a strong significant statistical difference with $p-value < .001$ (independent samples t-test).}
    \label{fig:exp_len}
\end{figure}

Since participants performed different numbers of moves, we aggregated those in slices (on the \textit{x-axis} in Figure \ref{fig:dt_com}) representing pieces of 5\% of the task to compare their moves' decision times. We then compared the participants' corresponding slices to have measures of similarity and easily visualize their moving times. However, we found that participants in the A-XAI group received less verbose explanations (of one word, on average) than those in the C-XAI group (independent samples t-test: $t = 2.31$, $p = .03$) as shown in Figure \ref{fig:exp_len}. We found no correlations between participants' personality traits, assessment behavioral measures, or persuasiveness. 

To summarize these findings, participants who interacted with the computer moved faster and were more resolute with the adaptive explanations than with the classical ones. This caused those who received the former to perform more actions than the other group of participants since the task was time-bounded.

\subsection{The influence of a humanoid social robot}
In this section, we show the results related to the Robot group. Contrary to the COM group, we found no behavioral differences between the two groups that interacted with the robot, neither in training nor in the assessment phase. We found that participants who interacted with the iCub robot performed a comparable number of actions during the training, regardless of the experimental group to which they belonged. Indeed, contrary to the COM group, we found no significant differences regarding the participants' decision time between the C-XAI and A-XAI groups. However, we found that participants in the A-XAI group received less verbose explanations (of one word, on average) than those in the C-XAI group (independent samples t-test: $t = 2.31$, $p = .03$) as shown in Figure \ref{fig:exp_len}.

As for the previous group, we found no correlations between the participants' personality traits and behavioral measures during training. However, regarding the assessment phase, we found that the amount of energy produced negatively correlated with participants' positive agency (Pearson's $r = -.455$, $p = .038$), and positively with their negative agency (Pearson's  $r = .484$, $p = .026$). Moreover, we found that also the number of anomalies - namely, conditions that damaged the NPP and restarted the application - negatively correlated with participants' positive agency (Pearson's $r = -.449$, $p = .041$).

\subsection{Comparisons between the computer and the humanoid robot}
In this section, we compare the results of the COM and Robot groups. To measure the influence that such artificial agents had on participants, we collected three types of moves for each step of the training phase:
\begin{itemize}
    \item \textit{Equal}: participants' originally selected action and the agents' suggested action were the same from the beginning.
    \item \textit{Follow self}: participants' originally selected action and the agents' suggested action differed, but they chose to confirm their initial indication.
    \item \textit{Follow AI}: participants' originally selected action and the agents' suggested action differed and they followed the agents' advice.
\end{itemize}

\begin{figure*}[t]
    \centering
    \includegraphics[width=.49\linewidth]{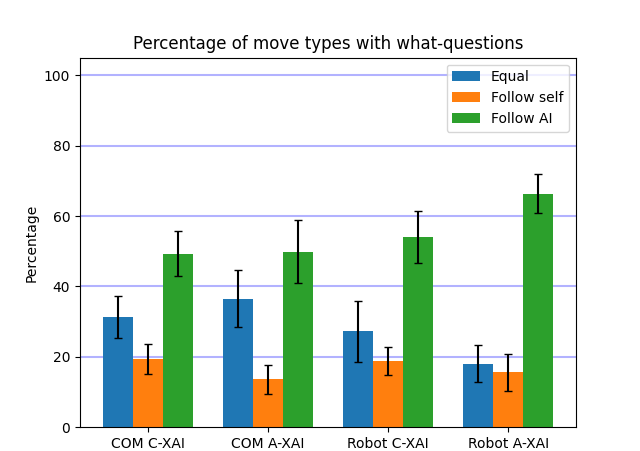}
    \includegraphics[width=.49\linewidth]{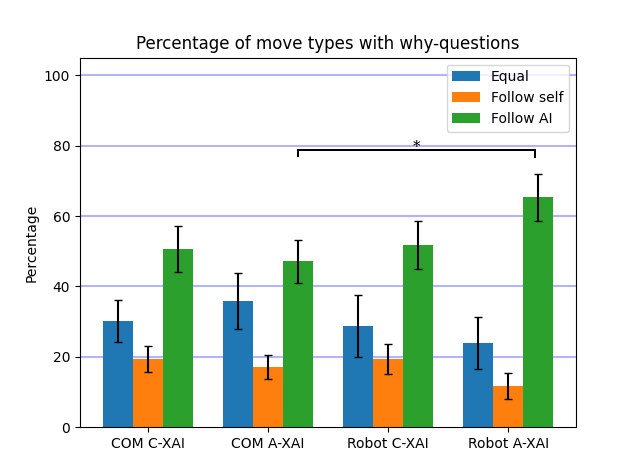}
    \caption{Average and std error of participants' move types during the training phase. The plot on the left refers to the step in which participants asked what questions, while the one on the right refers to the step in which they also asked why questions. It has to be noted that why questions always followed what questions. The difference reported with the * refers to a \textit{p-value} of .038 (independent samples t-test).}
    \label{fig:persuasiveness_npp}
\end{figure*}

To compare the move types of participants from different experimental groups, we averaged the percentages of their move types and tested those through independent samples t-tests. Regarding the what-questions, Figure \ref{fig:persuasiveness_npp} (left side), we found no significant differences in this regard. On the other hand, regarding the why-questions, Figure \ref{fig:persuasiveness_npp} (right side), we found a significant difference in the adaptive explanations between the COM and Robot group (independent samples t-test: $t = 2.226$, $p = .038$ ($\mu = 47.08$, $\sigma_M = 6.1$) for the COM group, and $t = 2.226$, $p = .038$ ($\mu = 65.37$, $\sigma_M = 6.68$) for the Robot group). Hence, the adaptive explanations made the robot more persuasive than the artificial agent when they justified their suggestions.

When interacting with the robot, the two explanation strategies did not cause any behavioral changes in the participants. However, the adaptive explanations resulted in a higher robot's persuasiveness. Hence, participants relied more on the robot's suggestions when justified with adaptive explanations than with classical ones. 

To summarize those findings, we can say that contrastive explanations automatized participants' behavior when interacting with the computer, making them move faster. On the other hand, they raised participants' tendency to follow the agent's suggestions when interacting with the social robot.


\begin{figure}[t]
    \centering
    \includegraphics[width=.8\linewidth]{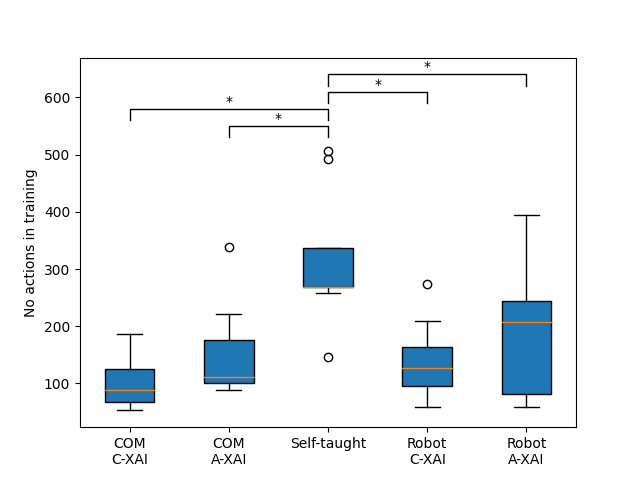}
    \caption{Number of actions performed during the training phase by all the groups. The * refers to strong statistical significance (ANOVA test with Bonferroni correction).}
    \label{fig:actions_all}
\end{figure}

\subsection{Comparison between self- and XAI assisted-learning}
\subsubsection{Behavioral measures}
We found an effect of the experimental condition on the number of actions performed in training among the experimental groups (ANOVA $F(4) = 8.66$, $p < .001$). Through a post hoc test with Bonferroni correction, we found that the Self-taught group performed more actions in training than all the other groups (Figure \ref{fig:actions_all}): $t = 3.654$, $p = .006$ with COM, A-XAI group; $t = 5.556$, $p < .001$ with COM, C-XAI group; $t = 3.24$, $p = .022$ with the Robot, A-XAI group; and $t = 4.459$, $p < .001$ with the Robot, C-XAI group. 

Similarly, there was an effect of the experimental condition on the energy produced during the training between the experimental groups (ANOVA $F(4) = 10.9$, $p < .001$). Through a post hoc test with Bonferroni correction, we found that the Self-taught group produced more energy than all the other groups:  $t = 4.684$, $p < .001$ with COM, A-XAI group; $t = 5.943$, $p < .001$ with COM, C-XAI group; $t = 3.992$, $p = .002$ with the Robot, A-XAI group; and $t = 5.344$, $p < .001$ with the Robot, C-XAI group. 

\begin{figure}[t]
    \centering
    \includegraphics[width=.8\linewidth]{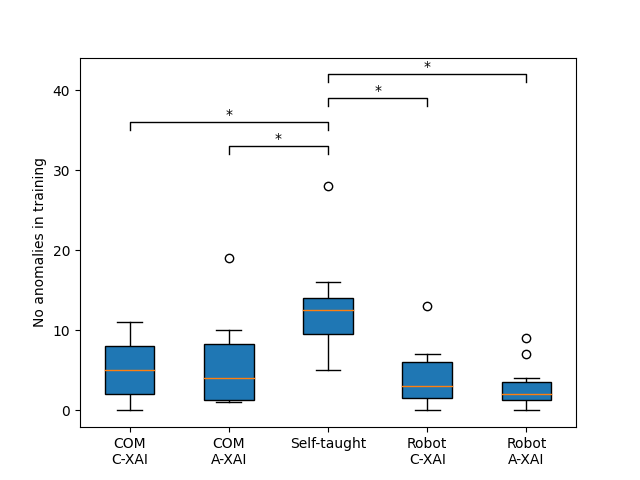}
    \caption{Number of anomalies produced during the training phase by all the groups. The * refers to strong statistical significance (ANOVA test with Bonferroni correction).}
    \label{fig:anomalies_all}
\end{figure}

We also found a statistically significant difference regarding the number of anomalies in training (ANOVA $F(4) = 6.86$, $p < .001$). Through a post hoc test with Bonferroni correction, we found that the Self-taught group produced more anomalies than all the other groups (Figure \ref{fig:anomalies_all}): $t = 3.345$, $p = .016$ with COM, A-XAI group; $t = 3.766$, $p = .005$ with COM, C-XAI group; $t = 4.255$, $p < .001$ with the Robot, A-XAI group; and $t = 4.683$, $p < .001$ with the Robot, C-XAI group. 

Finally, we found an effect of the experimental condition on the number of critic steps - namely, steps in which the NPP was actively producing energy - in training (ANOVA $F(4) = 6.52$, $p < .001$). Through a post hoc test with Bonferroni correction, we found that the Self-taught group performed more critic steps than the COM C-XAI group ($t = 3.293$, $p = .015$), the COM A-XAI group ($t = 4.817$, $p < .001$), and the Robot C-XAI one ($t = 3.765$, $p = .004$).

\begin{figure*}[t]
    \centering
    \includegraphics[width=.49\linewidth]{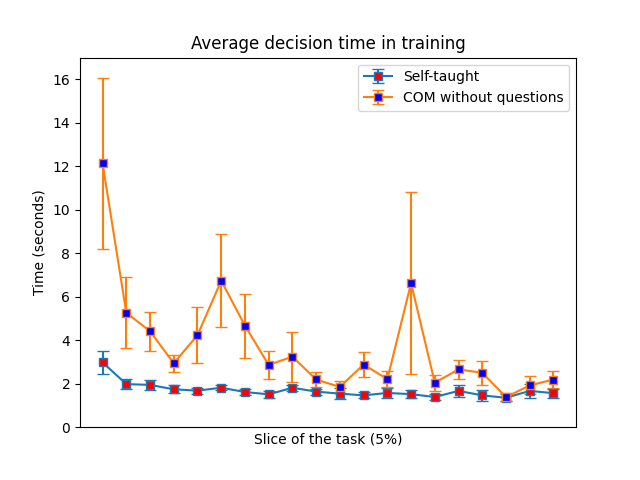}
    \includegraphics[width=.49\linewidth]{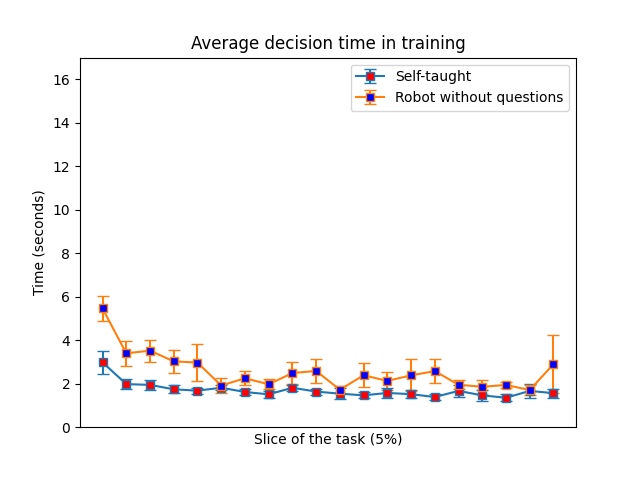}
    \caption{Average and std error of participants’ decision times during training. Each point of the x-axis represents 5\% of the training phase: it has to be read from the left to the right. The plot on the left compares the decision times of the Self-taught group and the COM group when they asked no questions. The plot on the right shows the comparison between the decision times of the Self-taught group and the Robot group when they asked no questions.}
    \label{fig:dt_self}
\end{figure*}

These results are explained by the shorter decision time participants in the Self-taught group had compared to the COM and Robot groups (Figure \ref{fig:dt_self}) (independent samples t-test: $t = -3.62$, $p < .001$, and $t = -4.12$, $p < .001$, respectively).

\subsubsection{Post-experiment test}
We found an effect of the experimental condition on the percentage of correct answers to the whole test when considering the Self-taught group (ANOVA $F(4) = 3.99$, $p = .007$). Through a post hoc test with Bonferroni correction, we found that participants belonging to the Self-taught group outperformed the COM C-XAI group ($t = 3.422$, $p = .013$), the COM A-XAI group ($t = 3.28$, $p = .02$), and the Robot A-XAI one ($t = 3.065$, $p = .036$) at the final test.

\begin{figure*}[t]
    \centering
    \includegraphics[width=.49\linewidth]{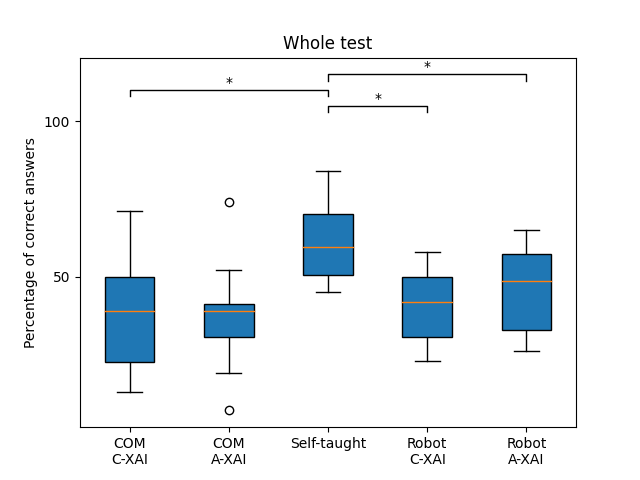}
    \includegraphics[width=.49\linewidth]{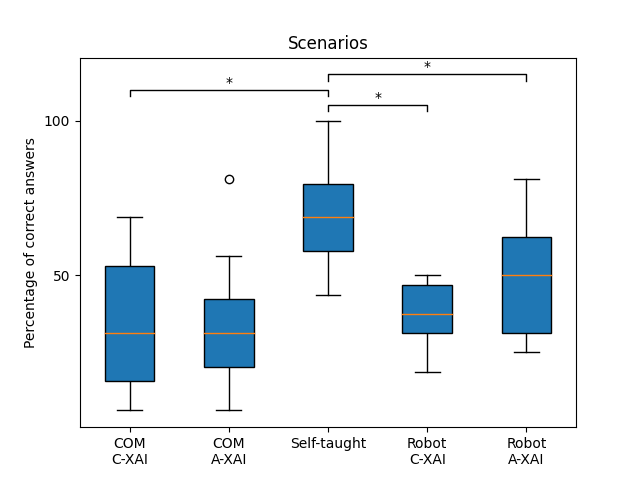}
    \caption{Distribution of the participants' correct answers to the whole test (left) and to the scenarios questions (right). The * refers to \textit{p-values} between .02 and .001.}
    \label{fig:boxplot_test}
\end{figure*}

None of the test sections showed significant differences between the conditions, except the part related to the \textit{scenarios} - namely, questions where participants were asked to report the causes of an indicated action, given precise conditions of the NPP - (ANOVA $F(4) = 6.37$, $p < .001$). Through a post hoc test with Bonferroni correction, we found that the Self-taught participants outperformed at the test's questions with the scenarios the COM C-XAI group ($t = 4.212$, $p = .001$), the COM A-XAI group ($t = 4.129$, $p = .001$), and the Robot A-XAI one ($t = 3.919$, $p = .003$). Figure \ref{fig:boxplot_test} shows the distributions of correct answers to the whole test (left side) and at the scenarios part (right side).

We found no interesting correlations between the Self-taught participants' behavioral measures and personality traits.

\section{Discussion}

Our study aimed to investigate the impact of explainable computers and humanoid robots on participants' behavior while learning an unknown task. Furthermore, we investigated whether contrastive (\textit{partner-aware}) explanations could elicit higher learning gain in such a context. Following our hypotheses, we expected that the partner-aware explanations would elicit more accurate participants' mental models about the task (H1), assisted participants (either by the computer or the robot) would produce comparable learning outcomes while outperforming the not-assisted ones (H2), and that the robot would influence participants' decision-making more than the computer (H3).

We analyzed participants' prior knowledge about nuclear power plants to ensure comparable initial conditions among the experimental groups. During the pre-experiment questionnaire, we asked participants to describe the functioning of a nuclear power plant (NPP) whose management represented the experimental task. Consequently, we coded such open-ended answers in three levels of knowledge and checked if there were differences in the distribution of the knowledge between the groups. Although such distributions were comparable, we could note a difference between those referring to the COM group and the other two: self-taught and robot. This slight difference can be explained by the origin of the participants belonging to the groups: those in the COM lived in Germany, while those in the other two lived in Italy. We hypothesized that German people have a higher awareness of nuclear energy because it has been used more recently than Italians; this brought slightly better knowledge about the functioning of NPPs. However, despite such an initial slight disparity, we registered no differences between these two macro-groups regarding performance in training and assessment.

Our between-subject factor was the explanation selection strategy. The artificial agents provided two kinds of explanations: classical (C-XAI) and adaptive (A-XAI). The first strategy selected the ``most important feature'' (in descending order) that the AI model used for its classifications, while the other one provided contrastive explanations based on participants' original action selection.

We checked whether those explanation selection strategies brought different occurrences of the explanandum (the topic of the explanation). We observed that this was the case for more or less half of the possible explanandum, especially for those with higher occurrences (e.g., temperature, pressure (Table \ref{tab:explanandums})). However, it has to be noted that the explanations, thus the explanandum, strictly depended on the participants' behavior during the learning.

Adaptive explanations caused participants who interacted with the COM computer to make faster decisions than those who received classical ones. We observed such an effect during the training and the assessment phases; thus, the explanation strategy affected participants' behavioral measures. Indeed, during training, the A-XAI group made faster decisions than the C-XAI one when they asked what- and why questions, but the decision time between the two groups was comparable when no question-asking was involved. For this reason, we can say that the difference in the moving time was mainly due to the explanation type that participants received and not to an individual difference in preferred decision speed. However, we found that classical explanations were more verbose than adaptive ones: this could have contributed to such differences. In our opinion, the difference in explanations' length was insufficient to explain the observed difference in decision timing, because the explanations differed only by one word on average: not even a second with the speed of speech used. Such limited difference does not seem to be  enough to claim that the explanations' verbosity was responsible for the difference in participants' decision times. 

Making faster decisions brought participants in the A-XAI group to perform more actions and produce more energy in training and assessment than in the C-XAI group. Despite those differences, the two explanation styles produced comparable exploration strategies in training (\textit{e.g.}, number of anomalies in training) and level of understanding of the task (\textit{e.g.}, results at the test), rejecting our hypothesis H1. Hence, we can say that the two explanation strategies brought only behavioral differences in the HCI context. Moreover, we found no effects of participants' personality traits on the interaction with the computer, partially confirming our hypothesis H4.

Conversely, regarding the Robot group, we found no behavioral differences brought by the two explanation strategies. Indeed, we observed a comparable number of actions performed, anomalies produced, and moving times between the C-XAI and A-XAI groups, both in training and assessment.

However, adaptive explanations influenced participants more with the humanoid robot than the computer. Indeed, the adaptive robot persuaded participants to opt for its actions (rather than their first selection) more than the computer did. Hence, we can say that adaptive explanations, on the one hand, empowered participants, making them more confident while interacting with the computer. On the other hand, they persuaded more participants who interacted with the humanoid robot. Nonetheless, the two explanation strategies proved to be similarly informative also for the Robot group, rejecting our hypothesis H1.

Contrary to our expectations, we found no correlations between the personality traits and behavioral measures of participants who interacted with the robot during training but only during the assessment phase. Thus, our hypothesis H3 has been partially confirmed. Indeed, the amount of energy produced in the assessment positively correlated with participants' negative agency and negatively correlated with their positive agency. Thus, the higher their negative agency and the lower their positive agency, the higher their performance in assessment. We speculate that the participants who tended to automatize their behavior - thus, unconsciously and unintentionally choosing their moves - had an advantage regarding their performance in assessment.
The further negative correlation between the number of anomalies in assessment and participants' positive agency supports our speculation about behavior automatism.

The correlations between participants' behavioral measures and move types tell us how the artificial agents influenced both the training and assessment. Regarding the training phase with both agents, we can see that the number of anomalies positively correlated with the percentage of moves in which participants maintained their original choices. Moreover, the former negatively correlated with the percentage of times participants accepted the robot's suggestions (and not the COM's).
It seems reasonable that the more seldom non-expert users follow the expert agent, the more errors they would make. 
However, the opposite (namely, the more participants followed the AI suggestions, the fewer errors they committed) turned out to be true only with the iCub robot. We speculate this was because of the higher persuasiveness of the robot, especially using contrastive explanations.

Regarding assessment, we can see negative correlations between participants' anomaly rate and the percentage of equal moves for both COM and Robot groups. On the other hand, we observed the opposite regarding participants' number of critic steps. It positively correlated with the equal moves for both groups. These results tell us that participants of both groups performed better in assessment when they could replicate the artificial agents' moves, that is, when they managed to learn during training from the choices and explanations by the agents.

As we expected, we observed that Self-taught participants performed more actions during training on average than those who interacted with both the artificial agents. This result is easily explainable by the greater availability of time they had compared to those participants who had to wait for the answers to their questions. However, Self-taught participants moved faster than the others, also when the latter did not ask any questions to the artificial agents. Hence, on equal terms, not-assisted participants made faster decisions than the assisted ones.

Consequently, we found that they produced more energy and performed more critic steps than those who interacted with the artificial agents. This is reasonable because we found that both those measures positively correlated with the number of actions performed. More interestingly, we observed that Self-taught participants produced more anomalies than the other groups.

In our opinion, the number of anomalies during training is a good estimator of participants' degree of exploration. Indeed, exploring the environment was crucial to understanding its functioning and achieving good test results. We observed that Self-taught participants outperformed those who interacted with the artificial agents at the post-experiment test, especially regarding the scenarios questions, rejecting our hypothesis H2. Those questions presented several environmental scenarios (with textual descriptions and pictures) and asked what would happen if one performed a specific action. We found significant differences between the Self-taught group and all the others but the COM A-XAI one. However, we can see that this was because of a participant in the COM A-XAI group who exhibited a substantially different behavior than all the others in their group since it is more than two standard deviations away from the mean (Figure \ref{fig:boxplot_test}). 

Since the number of anomalies in training is a good indicator of the exploration, we can say that Self-taught participants explored the environment more than the others. We speculate that this was so because no agent influenced them. Hence, they felt free to learn how the environment worked adequately without the pressure to avoid producing anomalies in front of an expert artificial agent. We can further hypothesize that self-taught participants did not suffer the harmful effect of the \textit{automation} biases \citep{vered2023effects} compared to those who interacted with the artificial agents. This is one of the most frequent cognitive biases - heuristics that the human brain produces to facilitate and speed up decision-making - in human-AI collaboration, by which people remain anchored to previous or others' ideas and suggestions.

Considering that we found no behavioral differences between Self-taught participants and the others regarding the assessment phase confirms our hypothesis that interacting with expert explainable artificial agents influenced the training but did not necessarily improve the training efficacy. Moreover, the moves in which participants changed their initial ideas to follow AI's suggestions were the most present move type in all conditions involving an interactive agent. We think participants in the COM and Robot groups limited themselves by asking many questions and following the agents' suggestions uncritically. Hence, we must reflect on how to deal with collaborative robots and AIs to ensure that people's learning is not limited since they seem to over-rely on such artificial agents.

We identified the reasons for the gap between the participants who interacted with the expert explainable artificial agents and those who did not interact with any of these interaction issues rooted in trust calibration, which have already been investigated, especially in the HCI context. Differently from \cite{naiseh2021explainable}, participants' lack of curiosity, perceived complexity, and perceived goal impediment (e.g., a too strong focus on the task completion) caused them to automatize their training and over-rely on the agents' suggestions. Furthermore, the reasons behind our results could be found in using over-simplified explanations, as observed in \cite{kulesza2013much}. While investigating how intelligent agents should explain to users, the authors observed that over-simplification can be a problem, causing participants to experience more mental demands. Subsequently, the additional cognitive load may have contributed to the emergence of heuristics about whether and how to follow the agents' suggestions. This problem has been recognized in the frame of the dual-process theory of cognition \cite{daniel2017thinking, kahneman2002representativeness, wason1974dual}, by which people rarely engage rationally with AI recommendations. 

Participants' over-confidence toward the explainable agents could also be due to the illusion of explanatory depth bias, which has already been observed regarding people's perceived understanding of additive local explanations \cite{chromik2021think}. Nonetheless, Bu\c{c}inca and colleagues individuated in \textit{cognitive forcing} and effective strategy to mitigate people's biases towards AI-generated suggestions \cite{bucinca2021trust}. They showed that cognitive forcing reduced over-reliance compared to simple XAI approaches, providing a starting point for mitigating people's bad habits in human-AI collaboration. However, to claim the intervention of cognitive biases in participants' decision-making needs deeper investigation, and we reserve the objective of investigating the occurrence of those biases in the future.

In our scenario, over-confidence toward the agents' suggestions seems an optimal strategy to maximize performance in training and assessment. However, our results demonstrate that it can bring people to limited capabilities to generalize, thus learning a new task effectively. Assisted participants exhibited good performance since the beginning because of the suggestions accepted. Hence, we want to highlight the trade-off between performing nicely in training, with low exploration, and learning to generalize. In our opinion, a smart AI assistant should avoid this effect by reducing peer pressure during training while inviting people to explore autonomously.

One limitation of our study regards the constraints we introduced in our experimental methodology. In particular, we refer to the fixed amount of time that participants had to perform the training phase. We designed it this way to introduce a time constraint. However, it is unclear whether we would observe the same results by removing or changing such a constraint, \textit{i.e.}, using a fixed number of steps. 
Another limitation regards the high variability of participants' behavior during both the training and assessment phases. Since they had no limitation in actions, participants showed high variability in performing their actions; thus, we could aggregate and analyze them in percentage.
Other limitations regard the specific robot platform used in our experiment, the iCub. Furthermore, the robot did not move too much and only expressed nice non-verbal behavior by making eye-contact while talking with participants and showing positive facial expressions. This aspect, together with its childish appearance, might have had an influence also on the generality of the results. 
The last limitation we would like to highlight is the use of self-reported questionnaires and possible biases that we might have inserted into the final test. To limit issues related to self-reported questionnaires, we used attention checks, and to reduce errors during the design of the test, we considered design principles from existing literature in multiple-choice question making.

\subsection{Conclusions}
In this study, we presented results from a user study in which participants had to perform a learning-by-doing task and interact with expert explainable artificial agents. In particular, we compared the effects of interacting with a computer and a social robot while performing our task. We compared two explanation strategies (classic versus partner-aware) using an assessment task to directly and quantitatively measure the informativeness of the agents' explanations. The assessment involved starting the task without knowledge, learning how to manage it by doing it, and completing a post-experiment test containing questions about the task itself. We tested a further group of participants who could not interact with any artificial agent, thus performing the task autonomously, and compared their behavior and knowledge with that of the XAI-assisted participants. Results showed how the two explanation strategies elicited no difference in task knowledge but produced different participants' behaviors depending on the agent providing them. Moreover, we highlighted that interacting with explainable agents can influence people's behavior and reduce their agency in learning-by-doing tasks, resulting in poor results compared to those of who learned autonomously. 

\section*{Acknowledgments}
This work has been supported by a Starting Grant from the European Research Council (ERC) under the European Union’s Horizon 2020 research and innovation programme. G.A. No 804388 (wHiSPER), and co-funded by the Deutsche Forschungsgemeinschaft (DFG): TRR 318/1 2021 – 438445824.

\section*{Declaration of Generative AI and AI-assisted technologies in the writing process}
While preparing this work, the authors used Grammarly and GPT 3.5 to check the grammar and improve the manuscript's readability. After using these tools, the authors reviewed and edited the content as needed and took full responsibility for the publication's content.


\end{document}